\pdfoutput=1

\documentclass[11pt]{article}
\usepackage[preprint]{acl}

\usepackage{times}
\usepackage{latexsym}

\usepackage[T1]{fontenc}
\usepackage[utf8]{inputenc}

\usepackage{microtype}
\usepackage{inconsolata}

\usepackage{graphicx}
\usepackage{amssymb}
\usepackage{pifont}
\newcommand{\cmark}{\ding{51}}
\newcommand{\xmark}{\ding{55}}

\usepackage{ragged2e}

\usepackage{amsmath,amsfonts,bm}

\def\eqref#1{equation~\ref{#1}}

\def\1{\bm{1}}

\DeclareMathAlphabet{\mathsfit}{\encodingdefault}{\sfdefault}{m}{sl}
\SetMathAlphabet{\mathsfit}{bold}{\encodingdefault}{\sfdefault}{bx}{n}

\usepackage{url}
\usepackage{graphicx}
\usepackage{subcaption}
\usepackage{multirow}
\usepackage{booktabs}
\usepackage[most]{tcolorbox}

\definecolor{block-gray}{gray}{0.85}
\newtcolorbox{blockquote}{colback=block-gray,grow to right by=-1mm,grow to left by=-1mm,boxrule=0pt,boxsep=0pt}

\title{In-Context Symbolic Regression: Leveraging Large Language Models for Function Discovery}

\author{Matteo Merler$^*$,~~ Katsiaryna Haitsiukevich$^*$,~~ Nicola Dainese$^*$ \and Pekka Marttinen\\
    Department of Computer Science \\ Aalto University \\ \texttt{\{firstname.lastname\}@aalto.fi}
    \\}

\begin{document}

\maketitle
\begin{abstract}
State of the art Symbolic Regression (SR) methods currently build specialized models, while the application of Large Language Models (LLMs) remains largely unexplored. In this work, we introduce the first comprehensive framework that utilizes LLMs for the task of SR.
We propose In-Context Symbolic Regression (ICSR), an SR method which iteratively refines a functional form with an LLM and determines its coefficients with an external optimizer. 
ICSR leverages LLMs' strong mathematical prior both to propose an initial set of possible functions given the observations and to refine them based on their errors.
Our findings reveal that LLMs are able to successfully find symbolic equations that fit the given data, matching or outperforming the overall performance of the best SR baselines on four popular benchmarks, while yielding simpler equations with better out of distribution generalization.
\end{abstract}

\def\thefootnote{*}\footnotetext{Denotes equal contribution.}\def\thefootnote{\arabic{footnote}}

\section{Introduction}
\label{sec:intr}

Classical Machine Learning regression methods can be divided into two broad categories: statistical methods, which learn an implicit statistical (black-box) model of the relationship between the observations, and rule-based methods, which instead attempt to extract an explainable set of rules that explicitly model the transformation between the inputs and outputs \citep{lample2019deep}. Symbolic Regression (SR) is a particular subset of the latter category, which searches the set of all possible explicit mathematical expressions to find the equation that best fits the given set of observations. This has the clear advantage of explainability, as well as a potential for better generalization, if the trend holds outside of the observed data.

The traditional approach for SR algorithms is Genetic Programming \citep{willis1997genetic} (GP), which combines fundamental blocks for mathematical expressions (e.g., basic operators, trigonometric functions, etc.) into more complex formulas using strategies borrowed from evolutionary biology, such as mutations and fitness. The recent success of Transformer models, first introduced by \citet{vaswani2017attention}, has revolutionized various fields of Artificial Intelligence, notably Natural Language Processing \citep{brown2020fewshot, openai2023gpt4, touvron2023Llama, anil2023palm} and Computer Vision \citep{dosovitskiy2021image}. Transformer-based methods have also been proposed for SR \cite{biggio2021neural, kamienny2022endtoend}, typically by employing a model pre-trained on a large amount of synthetic SR datasets.

Large Language Models (LLMs), also based on the Transformer, have proven to possess unprecedented reasoning and generalization abilities, based on their capability for In-Context Learning (ICL) \citep{brown2020fewshot}. This refers to the ability to perform tasks based on the context provided in the input text without any additional fine-tuning. With the help of ICL, these models can be leveraged for a wide range of different tasks, suggesting a potential use case for Symbolic Regression. 

\begin{figure*}
    \centering
    \includegraphics[width=\textwidth]{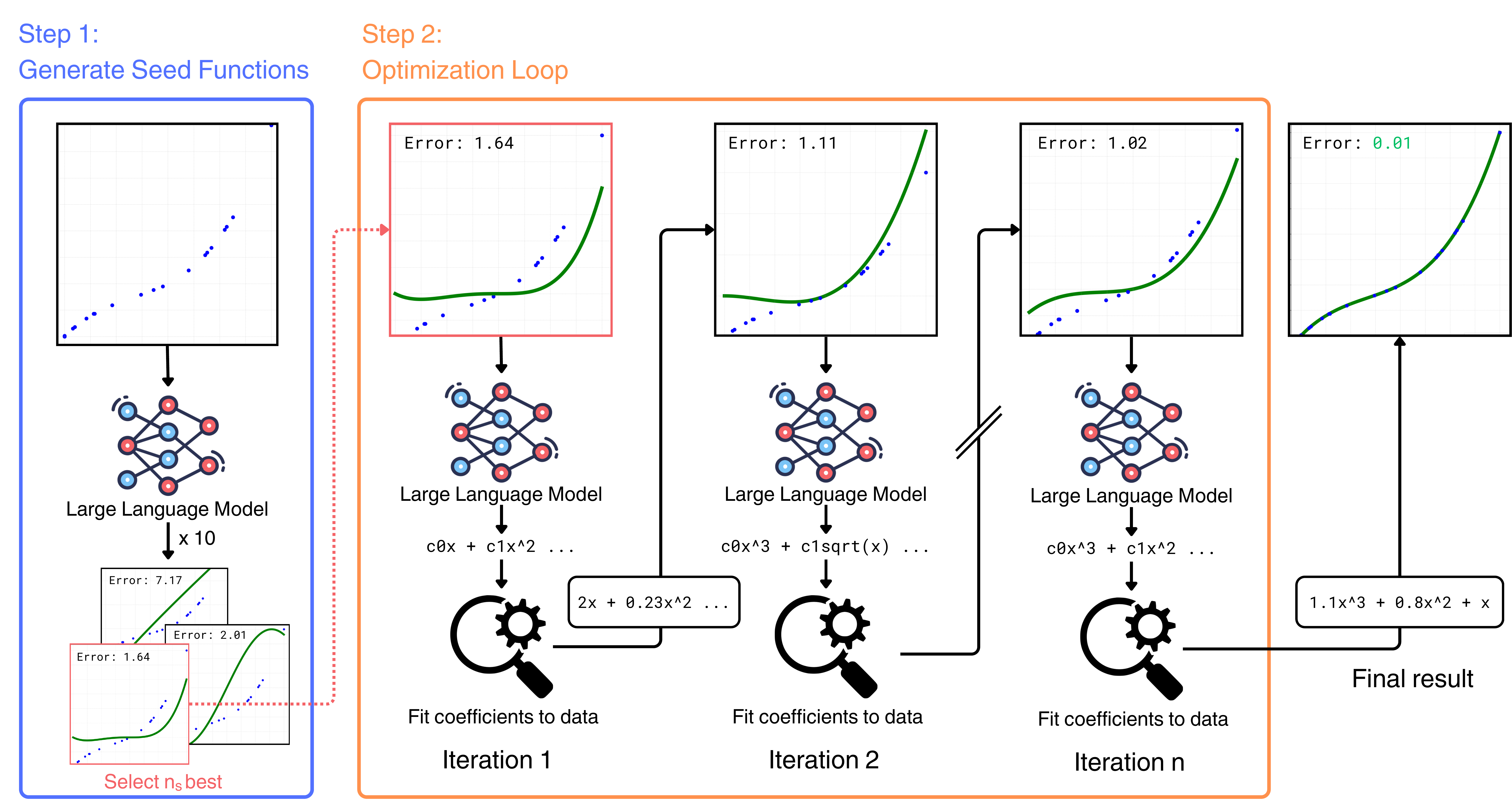}
    \caption{
    \textbf{High level overview of the ICSR approach.} Given an initial set of observations, we prompt the LLM to generate multiple initial guesses (seeds) of the true function that generated the observations.
    We then iteratively refine our guesses within an optimization loop where we propose new functions (based on a set of the previous best attempts), fit their coefficients and evaluate their fitness. The model only produces the functional form of a function, while the unknown coefficients are fitted using non-linear least squares optimization.
    }
    \label{fig:sketch}
\end{figure*}

In this paper, we examine the integration of LLMs into the SR pipeline, with the aim of using them to search for new equations that could fit the data. Inspired by the Optimization by Prompting (OPRO) approach presented by \citet{yang2023large}, we propose \textbf{In-Context Symbolic Regression (ICSR)}\footnote{We release the code at: \url{https://github.com/merlerm/In-Context-Symbolic-Regression}.}. This approach leverages pre-trained language models by providing a number of previously tested equations and their fitness scores in the prompt, tasking them to generate a new candidate that could be a better fit. The method is repeated until convergence is reached or the computational budget is exhausted.
To the best of our knowledge, only a contemporary work by \citet{shojaee2024llmsr} has ever explored the use of LLMs for SR. However, they focus on working with equations from a scientific domain where natural language knowledge can be directly incorporated, while this work aims to generally explore the capabilities of LLMs for SR without any additional information, in order to lay a foundation that can be expanded later. We discuss in depth the differences between the two works in Section~\ref{sec:rel}.

Our approach presents several advantages compared to models specifically trained for SR: as the LLM is not fine-tuned for this task, improvements in the underlying base model can improve ICSR without any changes to the method itself. Further, LLMs provide a natural language interface that can be leveraged to include additional information about the problem, like the domain of the equation and the interpretation of the observation values. The models could also be asked to explain the reasoning behind the proposed functions, potentially leading to a more interpretable process.

In summary, we make the following contributions: 
\textbf{1)} We propose ICSR, the first general framework to leverage LLMs for the SR task. 
\textbf{2)} We compare the method with a range of competitive SR baselines, matching or outperforming state of the art results on four popular SR benchmarks: Nguyen \citep{nguyen}, Constant \citep{li2023transformerbased}, R \citep{R} and Keijzer \citep{keijzerImprovingSymbolicRegression2003}.
\textbf{3)} We show that the equations generated with our method tend to exhibit lower complexity, which correlates with stronger out of distribution performance.

\section{Related Work}
\label{sec:rel}

\paragraph{Symbolic Regression.} GP has traditionally formed the backbone for SR methods \citep{Smits2005, Schmidt2011, Virgolin_2021}. Typically, from an initial population, an iterative tournament is played where functions with the highest fitness are selected to 'reproduce' with some random mutation, as in \citet{Koza2005}. 

More recently, Deep Learning methods have been applied to enhance the available toolkit for SR. \citet{udrescu2020ai} proposed an iterative simplification of the problem relying on insights from physics and outsourcing the function approximation part to a neural network. \citet{petersen2021deep} used a Recurrent Neural Network (RNN) with a risk-seeking policy to perform a hierarchical search over the space of user-defined operators and mathematical functions. The main drawback of these methods, including GP, is the fact that the algorithms start from scratch for every new expression, with very limited abilities of knowledge preservation between tasks.

To address this limitation, numerous Transformer-based methods inspired by language modelling have been developed. SymbolicGPT by \citet{valipour2021symbolicgpt}, NeSymReS by \citet{biggio2021neural} and CL-SR by \citet{li2023transformerbased} proposed different generative Transformer models specifically trained for SR. These models generate a functional form ('skeleton') of the equation with a special token for coefficients which are fitted via an external numerical optimizer. Subsequently, \citet{kamienny2022endtoend} presented E2E, a Transformer model able to produce the full expression including the coefficient values. While retaining knowledge between tasks, Transformer-based methods are quite limited in refining their solutions for the given set of points. 
To this end, \citet{shojaee2023transformerbased} presented a method integrating a pre-trained Transformer with Monte Carlo Tree Search to guide the equation generation merging the strength of the search and model pre-training. The proposed framework can be also viewed as a combination of a pre-trained model and an iterative refinement process.
However, none of the prior methods employ a foundation model \cite{bommasani2021opportunities}, such as an LLM, in order to leverage mathematical knowledge, but either pre-train an SR model \cite{biggio2021neural,kamienny2022endtoend}, or learn from scratch for every new function \cite{petersen2021deep}.

\paragraph{Mathematical Reasoning with LLMs.} As LLMs form the backbone of the method presented in this work, we rely entirely on their mathematical reasoning capabilities, such as ICL \citep{brown2020fewshot}, to explore the solution space.
\citet{mirchandani2023large} show that LLMs are able to recognize patterns from in-context examples and can extrapolate them to complete related tasks in the input. Similarly, \citet{gruver2023large} find that LLMs can extrapolate zero-shot the pattern from a timeseries (although they do not extract any functional representation). Furthermore, \citet{fu2023transformers} present a study in which they find that Transformer models can learn higher-order optimization methods (similar to Newton's method).

Contemporary to our work, \citet{shojaee2024llmsr} also propose to perform SR with an LLM aided by an external coefficient optimizer. However, they focus exclusively on the case where LLMs can leverage scientific knowledge for SR, by including a description of the input and output variables in the LLM prompt.
In contrast, we focus on the general case where no extra knowledge is given and test on standard benchmarks within the SR community and include a wider range of established baselines, with the aim to directly evaluate the capability of LLMs on the task of SR. Furthermore, we propose advancements in the structure of the prompt, including the coordinates of the points to be regressed, the score of previous attempts, and more in-context examples. 
Finally, rather than asking the LLM to optimize a Mean Squared Error (MSE) objective, we employ a more advanced loss function, presented in Section~\ref{sec:meth}, which jointly optimizes for the accuracy and complexity of the function for improved generalization properties.

\section{Background}
\label{sec:back}

The \textbf{Optimization by Prompting (OPRO)} framework was introduced by \citet{yang2023large} for prompt optimization, i.e., for increasing the performance of models (such as LLMs) that receive a textual prompt in the input and have to perform a specific task (such as mathematical reasoning).
Closer to our interest, the authors also present experiments on classical optimization problems (Linear Regression and Travelling Salesman Problem), suggesting that OPRO can solve such tasks. 

The key idea of the method is the use of a so-called meta-prompt, a higher level prompt which contains a description of the task to be optimized and previous attempts (examples) in solving it with their corresponding scores. An example of such task can be querying the model to find a linear function that fits a set of points. In this case, the prompt is augmented by the functions that have been tried out and the mean squared error on the data, obtained with an external evaluation procedure.
The assumption behind it is that LLMs have the ability to extrapolate the pattern formed by the examples, thanks to ICL, and propose a better alternative. 
The meta-prompt is given as input to the LLM and the model's output is then evaluated and added back to the meta-prompt if the score is good enough. This approach can then be iterated until a satisfying result is achieved or a certain computational budget is exhausted. 

\section{Method}
\label{sec:meth}
We consider a regression dataset $\mathcal{D}$ with $N$ observations and target variables $\{x_i,y_i\}_{i=1}^N$, also denoted with $(X,Y)$ more compactly, to be used for producing a function $\hat{f}$ that well approximates the data.
To leverage the OPRO approach for SR, we need to design a meta-prompt suitable for the task and fill it with the available observations $(X,Y)$, an initial set of $k$ functions $\hat{F}_{0} = \{\hat{f}_0^{(1)}, \hat{f}_0^{(2)}, \dots, \hat{f}_0^{(k)}\}$ (either hand-written or model-generated) and a measure of their fitness (score) on $\mathcal{D}$. For our purposes, we frame the refinement process as a minimization problem over an objective function, also called the error function, such that generated equations with the lowest error have the highest fitness. The goal is then to iteratively refine the set of functions $\hat{F}_{i}, i \in [1, \dots, n]$ for each iteration $i$, until a sufficiently low error is obtained by one of them or a maximum number of iterations is reached; we denote this process as the \textbf{optimization loop}. Due to the finite size of the LLM context window, we only keep the $k$ best performing previous attempts in the set $\hat{F}_i$, where $k$ is a design choice ($k=5$ in this study). The meta-prompt used in the experiment can be found in Appendix~\ref{app:prompts}.

\paragraph{Seed Functions.} At the first iteration, $\hat{F}_0$ is empty as there are no previous guesses from the model. Thus, an initial population of seed functions is required to kickstart the optimization loop.
Instead of relying on a fixed set of initial functions, which could be restrictive in general, we ask the model to generate the initial seed functions (with the prompt provided in Appendix~\ref{app:prompts}). This results in a complex and diverse set of functions, from which the LLM can refine its future predictions with the optimization loop. In our implementation we repeat this initial process $n_s$ times, as some of the generated functions can be undefined for certain input points (e.g., $\log(x)$ for negative numbers). 
We set $n_s=10$ for this work and explore its impact in Section~\ref{sec:exp:abl} through an ablation study.

\paragraph{Error Function.} The immediate choice for the objective function would be an MSE, or a similar error metric, over the regression dataset $\mathcal{D}$. However, simply minimizing this error can result in overfitting on the training points in $\mathcal{D}$. As overfitting in SR often occurs due to a growing number of terms in the generated equation, we adapt from \citet{shojaee2023transformerbased} a fitness function $r(\hat{f}|\mathcal{D})$ with an extra penalty term for the complexity $\mathcal{C}$ of the generated expression, defined as the following:
\begin{equation}
    r(\hat{f}|\mathcal{D}) = \frac{1}{1 + \text{NMSE}(\hat{f}|\mathcal{D})} + \lambda e^{\left(-\frac{\mathcal{C}(\hat{f})}{L} \right)},
\end{equation}
where $\hat{f}$ is the predicted function, $\mathcal{C}$ is the complexity defined as the number of nodes in the expression tree, $L$ is the maximum sequence length (set to 30), and $\lambda$ is a hyperparameter to trade-off between the fit to the data and the complexity. The Normalized Mean Square Error (NMSE) is calculated as 
\begin{equation}
\text{NMSE}(\hat{f}|\mathcal{D}) = \frac{\sum_{i=1}^N(y_i-\hat{f}(x_i))^2}{\sum_{i=1}^N y_i^2 + \epsilon}\,,
\end{equation}
where $\epsilon$ is a small regularizing constant.
Finally, we use $\text{err}(\hat{f}|\mathcal{D}) = r(\hat{f}|\mathcal{D})^{-1}$ as our error function to frame ICSR as a minimization problem. We explore the choice of the $\lambda$ parameter in Section~\ref{sec:exp:abl} with a sensitivity analysis.

\paragraph{Parameter Fitting.}
We utilize the LLM only to generate functional forms (skeletons), while the unknown coefficients associated to the predicted functional form are optimized by Non-linear Least Squares (NLS) \citep{kelley1999iterative} available from SciPy's \citep{2020SciPy-NMeth}. 
This not only yields better coefficient values, due to the superior optimization performance of NLS over LLMs, but also allows for more efficient exploration of the space of functions, by grouping them in equivalence classes of unique functional forms.
In our implementation, we optimize the function's coefficients five times starting from different random initial values, to avoid local minima, similarly to \citet{li2023transformerbased}.
\paragraph{}
For other details about the OPRO implementation, we follow the original work. Specifically, we also sample multiple functions for every iteration (asking the model to generate 5 functions for every call) in an attempt to improve the stability of the loop and we experiment with a decreasing temperature parameter to balance exploration/exploitation (with a higher initial temperature encouraging the exploration of the underlying functional space, and a lower temperature at the later stages forcing smaller tweaks to the trajectory). To avoid saturating the model's context window, we limit the amount of training points that are included in written form to a certain threshold, empirically set to 40. We discuss this in more detail in the Limitations Section~\ref{sec:disc:limit}.

\begin{table*}[h!]
    \centering
    \resizebox{\linewidth}{!}{
    \begin{tabular}{lccccccc}
        \toprule
         \multirow{2}{*}{\textbf{Method}} & \textbf{SR training} & \textbf{Evaluated} & \textbf{Model} & \textbf{Pre-trained} & \textbf{Flexible} & \textbf{Problem specific} & \textbf{Complexity}\\ 
          & \textbf{examples} & \textbf{expressions}  & \textbf{size} & \textbf{model} &\textbf{vocabulary} & \textbf{refinement} & \textbf{penalty}\\
          \midrule
          ICSR (Ours) & 0 & $\mathcal{O}((50\cdot5 + 10)\cdot5)$ & 8B & \cmark & \cmark & \cmark & \cmark \\
          \hline
          gplearn     & 0 & $\mathcal{O}(1000\cdot20)$ & - & \xmark & \xmark & \cmark & \xmark \\
          DSR         & 0 & $\mathcal{O}(200\text{K})$ & 8K & \xmark & \xmark & \cmark & \xmark \\
          uDSR*       & 0 & $\mathcal{O}(200\text{K})$ & 8K & \cmark / \xmark & \xmark & \cmark & \xmark \\
          NeSymReS    & 100M & $\mathcal{O}{(10\cdot10)}$ & 26M & \cmark & \xmark & \xmark & \xmark \\
          E2E         & 3M & $\mathcal{O}(100)$ & 86M & \cmark & \xmark & \xmark & \xmark \\
          TPSR        & 3M & $\mathcal{O}(200\cdot3)$ & 86M & \cmark & \xmark & \cmark & \cmark \\
          \bottomrule
          \multicolumn{8}{l}{* The uDSR method potentially allows using a pre-trained model as a prior. However, as reported in the original paper,} \\
          \multicolumn{8}{l}{while this is useful in a low-budget search it has tendencies to worsen the performance.}
    \end{tabular}
    }
    \caption{\textbf{Qualitative comparison across baselines.} We compare different properties for all baselines. \textbf{Evaluated expressions} is the total number of equations a method considers for modeling a given training set. \textbf{Pre-trained model} refers to the use of an underlying model as opposed to training from scratch for each problem. \textbf{Problem specific refinement} refers to the use of a search algorithm on the space of possible skeletons.}
    \label{tab:benchmarks_q}
\end{table*}

\section{Experiments}
\label{sec:exp}
We empirically evaluate ICSR and compare it against a set of competitive baselines, checking both in-domain and out of distribution performance of the proposed approach.

\subsection{Benchmarks}
\label{sec:exp:bench}
For our experiments, we choose four popular SR benchmarks containing functions with one or two input dimensions: \textbf{Nguyen} \citep{nguyen}, \textbf{Constant} (a modified version of some of the Nguyen equations with different numerical values for the coefficients \citep{li2023transformerbased}), \textbf{R} \citep{R} and \textbf{Keijzer} \citep{keijzerImprovingSymbolicRegression2003}.
The symbolic equations and ranges for both the training and testing points are reported in Appendix~\ref{app:bench_func}. We leave for future work the evaluation of ICSR on higher dimensionality benchmarks.

\begin{table*}[h!]
    \centering
    \resizebox{\textwidth}{!}{
    \begin{tabular}{lcccccccc|cc}
         \toprule
         \multirow{2}{*}{\textbf{Method}} & \multicolumn{2}{c}{\textbf{Nguyen} ($\bar{\mathcal{C}}=5.2$)} & \multicolumn{2}{c}{\textbf{Constant} ($\bar{\mathcal{C}}=4.3$)} & \multicolumn{2}{c}{\textbf{R} ($\bar{\mathcal{C}}=8.3$)} & \multicolumn{2}{c}{\textbf{Keijzer} ($\bar{\mathcal{C}}=5.0$)} & \multicolumn{2}{c}{\textbf{Overall avg.}} \\
         \cmidrule{2-11}
          & $R^2$ ($\uparrow$) & $\mathcal{C}$ ($\downarrow$) & $R^2$ ($\uparrow$) & $\mathcal{C}$ ($\downarrow$) & $R^2$ ($\uparrow$) & $\mathcal{C}$ ($\downarrow$) & $R^2$ ($\uparrow$) & $\mathcal{C}$ ($\downarrow$) & $R^2$ ($\uparrow$) & $\mathcal{C}$ ($\downarrow$) \\    
          \midrule
          ICSR (Ours) & 0.996 $\pm$ 0.002 & 6.4 $\pm$ 0.5 & 0.9991 $\pm$ 0.0004 & \textbf{4.6 $\pm$ 0.4} & \textbf{0.996 $\pm$ 0.001} & 7.5 $\pm$ 0.3 & \textbf{0.981 $\pm$ 0.004} & 8.2 $\pm$ 0.7 & \textbf{0.993 $\pm$ 0.002} & 6.7 $\pm$ 0.4 \\
          \hline
          gplearn     & 0.75 $\pm$ 0.15 & 7.2 $\pm$ 0.8 & 0.74 $\pm$ 0.22 & 5.6 $\pm$ 0.7 & 0.97 $\pm$ 0.01 & 7.6 $\pm$ 1.3 & 0.09 $\pm$ 0.43 & 10.5 $\pm$ 1.4 & 0.6 $\pm$ 0.2 & 7.7 $\pm$ 1.0 \\
          DSR         & 0.983 $\pm$ 0.005 & \textbf{5.8 $\pm$ 0.3} & 0.96 $\pm$ 0.01 & 6.9 $\pm$ 0.7 & 0.95 $\pm$ 0.03 & \textbf{5.7 $\pm$ 0.5} & 0.84 $\pm$ 0.03 & 6.3 $\pm$ 0.3 & 0.93 $\pm$ 0.02 & \textbf{6.2 $\pm$ 0.5} \\
          uDSR        & \textbf{0.9998 $\pm$ 0.0001} & 20.4 $\pm$ 1.1 & \textbf{0.9997 $\pm$ 0.0001} & 21.9 $\pm$ 1.5 & 0.993 $\pm$ 0.004 & 15.3 $\pm$ 0.5 & 0.980 $\pm$ 0.005 & 22.4 $\pm$ 1.5 & \textbf{0.993 $\pm$ 0.002} & 20.0 $\pm$ 1.2	 \\
          NeSymReS    & 0.976 $\pm$ 0.007 & 6.3 $\pm$ 0.2 & 0.97 $\pm$ 0.01 & 5.9 $\pm$ 0.2 & 0.92 $\pm$ 0.02 & 6.2 $\pm$ 0.5 & 0.87 $\pm$ 0.02 & \textbf{6.2 $\pm$ 0.2} & 0.93 $\pm$ 0.01	& \textbf{6.2 $\pm$ 0.3} \\
          E2E         & 0.9976 $\pm$ 0.0005 & 18.1 $\pm$ 1.1 & 0.996 $\pm$ 0.002 & 16.8 $\pm$ 1.2 & 0.68 $\pm$ 0.20 & 22.3 $\pm$ 1.3 & 0.82 $\pm$ 0.05 & 20.2 $\pm$ 0.9 & 0.87 $\pm$ 0.06	& 19.4 $\pm$ 1.1 \\
          TPSR        & \textbf{0.9998 $\pm$ 0.0001} & 13.7 $\pm$ 0.6 & 0.9993 $\pm$ 0.0001 & 11.5 $\pm$ 0.7 & \textbf{0.996 $\pm$ 0.001} & 13.3 $\pm$ 0.7 & 0.92 $\pm$ 0.03 & 17.2 $\pm$ 0.8 & 0.979 $\pm$ 0.008 & 14.0 $\pm$ 0.7 \\
          \bottomrule
    \end{tabular}
    }
    \caption{\textbf{Comparison across baselines.} We evaluate each method on all benchmarks with five random seeds, reporting the averages for the coefficient of determination $R^2$ and the function complexity $\mathcal{C}$ with the error of the mean.
    We further report the average ground truth complexity for each benchmark, indicated with $\bar{\mathcal{C}}$.}
    \label{tab:benchmarks}
\end{table*}

\subsubsection{Metrics}
\label{sec:exp:bench:metrics}

While we use the error function $\text{err}(\hat{f}|\mathcal{D})$ during the optimization loop (see Section~\ref{sec:meth}), we follow the literature in reporting the coefficient of determination $R^2$ \citep{r2} to evaluate the quality of our method. This staple metric in SR can be interpreted as follows: a function will get a positive score if it is more accurate than the average prediction and will get a score of 1 for a perfect prediction. The coefficient is computed as:
\begin{equation}
R^2 = 1 - \frac{\sum_{i=1}^n (y_i - \hat{y}_i)^2}{\sum_{i=1}^n (y_i - \bar{y})^2},
\end{equation}
where $y_i$ is the ground truth value, $\hat{y}_i$ is the predicted value and $\bar{y}$ is the average of all $y_i$.
 
We report the $R^2$ metric computed on a set of unseen testing points, obtained from a dense grid within the same range of input point values as during training (for in-domain performance) or its extended version (for out of distribution performance).
We follow \citet{li2023transformerbased} and \citet{biggio2021neural} in removing the 5\% worst predictions in all methods to ensure robustness against outliers. We further report the complexity $\mathcal{C}$ of the generated equations, calculated as the number of nodes in its expression tree.
For all methods, we repeat all experiments across five different random seeds and report the average values together with the standard error of the mean. For ICSR, we allow up to 50 iterations in the optimization loop and end it earlier if the $R^2$ score on the training set exceeds 0.99999.

\subsection{Baselines}
To evaluate the performance of the proposed method we opted for the following list of competitive baselines: \textbf{gplearn} \cite{gplearn}, a classical GP approach; \textbf{DSR} \cite{glatt2022deep} and \textbf{uDSR} \cite{landajuela2022uDSR}, two search-based methods; \textbf{NeSymReS} \cite{biggio2021neural} and \textbf{E2E} \cite{kamienny2022endtoend}, selected as representatives for Transformer-based model pre-trained over a large-scale SR dataset; and \textbf{TPSR} \cite{shojaee2023transformerbased}, which augments E2E with a decoding strategy guided by Monte-Carlo Tree Search, as an efficient combination of pre-training and search.
The details of the baseline model and the hyperparameters can be found in Appendix~\ref{app:hyper}.

We compare various properties of the considered methods in Table~\ref{tab:benchmarks_q}. Thanks to the use of LLMs, ICSR is able to leverage a much larger model size without the need for SR-specific training examples, as opposed to the other Transformer based methods. Furthermore, our method is far more sample efficient when compared to search-based methods like DSR and uDSR. The LLM is slower in generating a single expression, but is able to produce more meaningful equations, thanks to the large pre-training bias, as opposed to methods like gplearn and DSR which have to be trained from scratch on each problem. ICSR is also the only method with a natural language interface and a flexible vocabulary, which we discuss further in Section~\ref{sec:disc}.

\subsection{Comparison across Baselines} 
\label{sec:exp:perf}
For comparison of ICSR with the baselines, we choose Llama 3 8B \citep{llama3} as the underlying LLM.
The results (see Table~\ref{tab:benchmarks}) show that
the ICSR approach is very robust, consistently achieving very high scores across all benchmarks while producing expressions with a lower average complexity. The overall average columns show that ICSR outperforms all baselines, with only uDSR matching its $R^2$ score at the cost of significantly higher complexity. In general, it is important to consider both metrics simultaneously, as simpler functions can lead to a sightly lower $R^2$ value while bringing other advantages, such as better out of distribution generalization, which we explore in Section~\ref{sec:exp:ood}. As seen in the headers in Table~\ref{tab:benchmarks}, the complexity values of the ground truth equations align much more closely to the ones recovered by ICSR as opposed to the ones for other high-performing baselines, such as uDSR or TPSR.
The improvement in complexity compared to TPSR is particularly noteworthy, as both methods are using the same objective function: this could be a sign that LLMs tend to produce more human-readable expressions thanks to their pre-training bias. It is also worth noting that ICSR can potentially improve over time by simply increasing the performance of the underlying LLM backbone without any additional training, while that is not the case for the other methods.

\begin{figure}[!h]
    \centering
    \includegraphics[width=1.0\linewidth,trim={11mm 2mm 15mm 2mm},clip]{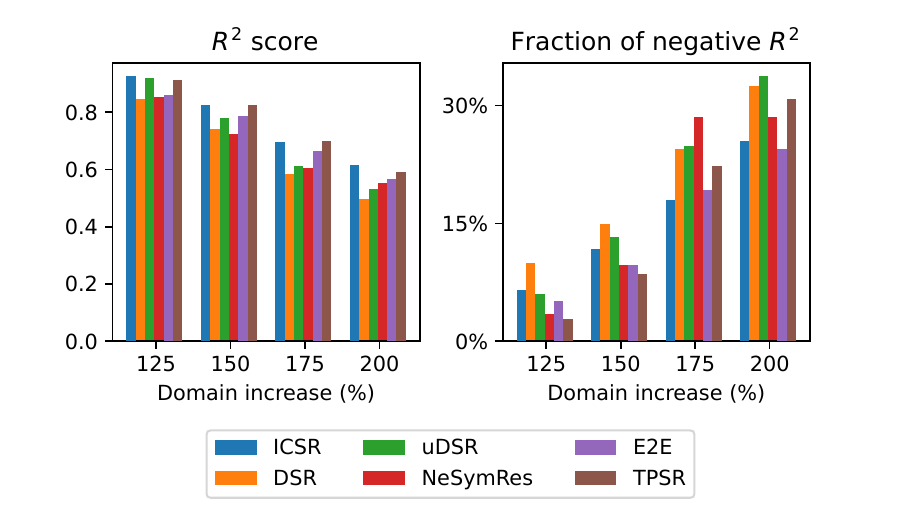}
    \caption{\textbf{Comparison across baselines on out of distribution data.} We compared the proposed method with the baselines by increasing the input domain for the generated functions. Whenever the $R^2$ becomes negative, we fix it to 0 when computing the average for the figure on the left and report the fraction of negative values in the figure on the right.}
    \label{fig:extrap}
\end{figure}

\begin{figure}[!h]
    \centering
    \begin{subfigure}{0.48\linewidth}
        \includegraphics[width=1.0\textwidth,keepaspectratio]{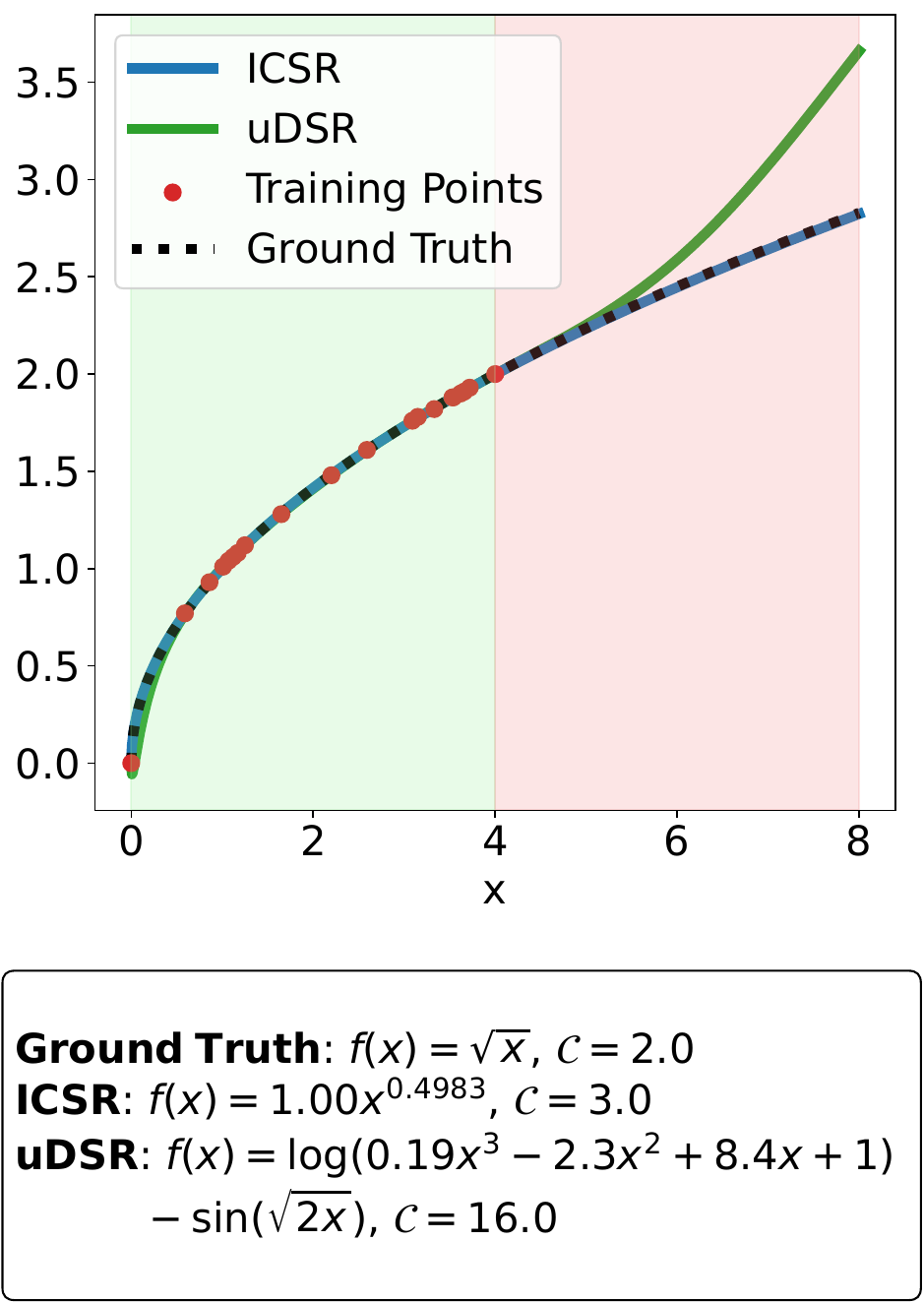} 
        \caption{Nguyen 8.}
    \end{subfigure} \begin{subfigure}{0.48\linewidth}
        \includegraphics[width=1.0\textwidth,keepaspectratio]{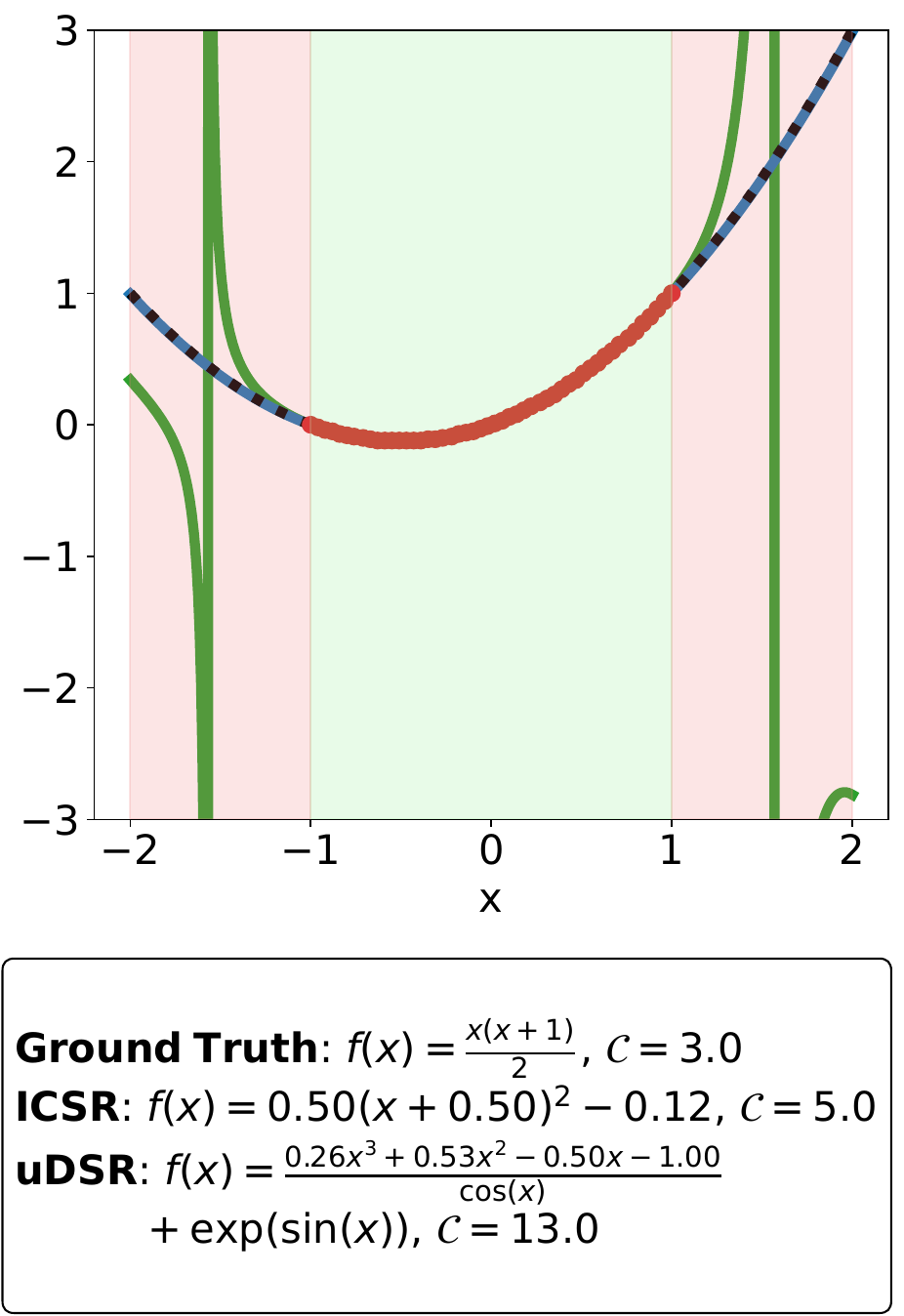}
        \caption{Keijzer 6.}
    \end{subfigure}
    \caption{\textbf{Out of distribution examples.} Qualitative examples demonstrating the generalization capabilities of ICSR and uDSR on two experiments. The higher complexity from the uDSR examples introduces unnecessary terms that harm the out of distribution performance (area shaded in red).}
    \label{fig:ood_examples}
\end{figure}

\begin{table*}[h!]
    \centering
    \resizebox{\textwidth}{!}{
    \begin{tabular}{lcccccccc}
         \toprule
         \multirow{2}{*}{\textbf{Method}} & \multicolumn{2}{c}{\textbf{Nguyen}} & \multicolumn{2}{c}{\textbf{Constant}} & \multicolumn{2}{c}{\textbf{R}} & \multicolumn{2}{c}{\textbf{Keijzer}} \\
         \cmidrule{2-9}
          & $R^2$ ($\uparrow$) & $\mathcal{C}$ ($\downarrow$) & $R^2$ ($\uparrow$) & $\mathcal{C}$ ($\downarrow$) & $R^2$ ($\uparrow$) & $\mathcal{C}$ ($\downarrow$) & $R^2$ ($\uparrow$) & $\mathcal{C}$ ($\downarrow$) \\
          \midrule
          ICSR ($\lambda=0.05$) & 0.996 $\pm$ 0.002 & 6.4 $\pm$ 0.5 & 0.9991 $\pm$ 0.0004 & 4.6 $\pm$ 0.4 & \textbf{0.996 $\pm$ 0.001} & 7.5 $\pm$ 0.3 & \textbf{0.981 $\pm$ 0.004} & 8.2 $\pm$ 0.7 \\
          \midrule
          ICSR ($\lambda=0$) & \textbf{0.9990 $\pm$ 0.0004} & 18.6 $\pm$ 0.5 & \textbf{0.9994 $\pm$ 0.0001} & 17.6 $\pm$ 0.6 & 0.988 $\pm$ 0.005 & 19.9 $\pm$ 0.8 & 0.92 $\pm$ 0.05 & 16.6 $\pm$ 0.6 \\
          ICSR ($\lambda=0.1$) & 0.992 $\pm$ 0.003 & 6.1 $\pm$ 0.5 & 0.9978 $\pm$ 0.0006 & 4.3 $\pm$ 0.3 & 0.989 $\pm$ 0.004 & 6.5 $\pm$ 0.4 & 0.980 $\pm$ 0.005 & 7.8 $\pm$ 0.6 \\
          ICSR ($\lambda=0.5$) & 0.94 $\pm$ 0.03 & 4.4 $\pm$ 0.4 & 0.983 $\pm$ 0.003 & 3.0 $\pm$ 0.2 & 0.972 $\pm$ 0.005 & 4.07 $\pm$ 0.07 & 0.95 $\pm$ 0.02 & 6.4 $\pm$ 0.6 \\
          ICSR ($\lambda=1$) & 0.92 $\pm$ 0.03 & \textbf{3.7 $\pm$ 0.3} & 0.89 $\pm$ 0.04 & \textbf{2.5 $\pm$ 0.2} & 0.95 $\pm$ 0.01 & \textbf{3.4 $\pm$ 0.2} & 0.77 $\pm$ 0.05 & \textbf{4.8 $\pm$ 0.5} \\
          \midrule
          Seed only ($n_s=10$) & 0.95 $\pm$ 0.03 & 11.4 $\pm$ 0.8 & 0.982 $\pm$ 0.003 & 8.0 $\pm$ 0.7 & 0.990 $\pm$ 0.005 & 10.3 $\pm$ 0.5 & 0.93 $\pm$ 0.03 & 10.8 $\pm$ 0.8 \\
          Seed only ($n_s=5$) & 0.986 $\pm$ 0.003 & 12.4 $\pm$ 0.7 & 0.986 $\pm$ 0.003 & 10.0 $\pm$ 0.7 & 0.995 $\pm$ 0.001 & 10.7 $\pm$ 0.6 & 0.88 $\pm$ 0.04 & 12.7 $\pm$ 0.8 \\
          Seed only ($n_s=1$) & 0.91* $\pm$ 0.04 & 14.5* $\pm$ 0.8 & 0.95* $\pm$ 0.03 & 13.5* $\pm$ 0.9 & 0.97* $\pm$ 0.02 & 12.8* $\pm$ 0.7 & 0.66* $\pm$ 0.07 & 16.0* $\pm$ 0.7 \\
          \midrule
          Random Guessing & 0.960 $\pm$ 0.006 & 3.9 $\pm$ 0.2 & 0.971 $\pm$ 0.005 & 4.4 $\pm$ 0.2 & 0.91 $\pm$ 0.03 & 4.7 $\pm$ 0.2 & 0.77 $\pm$ 0.04 & 4.0 $\pm$ 0.2 \\
          \bottomrule
          \multicolumn{9}{l}{* Some runs failed to generate valid seed functions. Only 88\% of the experiments for nguyen, 93\% for constant, 87\% for R and 95\% for keijzer finished} \\
          \multicolumn{9}{l}{with at least one valid function.} \\
    \end{tabular}
    }
    \caption{\textbf{Sensitivity analysis and ablation studies.} We perform sensitivity analysis on the values of the complexity penalty parameter $\lambda$ and two ablation studies: one using only $n_s$ initial seed functions without improving them and the other one using random guessing, rather than ICSR, for proposing new functions. All ablations on $n_s$ are performed without the optimization loop, only keeping the best generated seed function. We report the averages for the coefficient of determination $R^2$ and the function complexity $\mathcal{C}$ with the error of the mean for all experiments. We highlight in \textbf{bold} the best performance across different values of $\lambda$.
    }
    \label{tab:ablation}
\end{table*}

\subsection{Out of Distribution Performance}
\label{sec:exp:ood}

We further explore the advantage of producing functions with a lower complexity value by testing the out of distribution capabilities of the expressions recovered by ICSR and the other baselines. We exclude gplearn from these experiments, as we observe its performance to be significantly lower compared to the rest of the methods. To include out-of-domain test points, we extend the input range by 100\% to all directions (in which the function is defined). We compute the $R^2$ value on the extended range, reporting the results in Figure~\ref{fig:extrap}. Note that the $R^2$ value can quickly become increasingly negative when the functions diverge significantly. In order to keep the results stable, we treat all negative values as 0 when computing the average and report the fraction of experiments with a negative $R^2$.

In general, we observe a sharp decline in performance for all methods, with the fraction of negative $R^2$ values quickly increasing towards the further extensions of the range. Specifically, ICSR is the highest performing method in the 175\% and 200\% domain increases, with 
the second lowest and lowest number of failures respectively. Generally, methods with lower complexity such as NeSymReS, E2E and TPSR tend to perform better than uDSR, with the exception of DSR which exhibits the poorest out of distribution performance even with a low average complexity.
The comparison between ICSR and uDSR is particularly meaningful: as reported in Table~\ref{tab:benchmarks}, the two methods are tied for the best overall average performance, but ICSR outperforms uDSR when extrapolating further outside of the training range thanks to the lower complexity of the recovered expressions. We present some qualitative examples that demonstrate the difference between the methods in Figure~\ref{fig:ood_examples}.

\subsection{Sensitivity Analysis and Ablation Studies}
\label{sec:exp:abl}

In this section we first investigate the impact of the $\lambda$ parameter and then test the importance of the iterative refinement of the equations with the optimization loop. Finally, we compare ICSR with a baseline where the LLM was not given any information about the observations. All results are reported in Table~\ref{tab:ablation}.

\paragraph{Lambda Parameter.} In our sensitivity analysis we considered the complexity penalty parameter $\lambda=[0, 0.05, 0.1, 0.5, 1]$. We noticed that the smallest penalty $\lambda=0.05$ was already sufficient to considerably reduce the complexity of the selected functions and increasing the penalty further had a relatively smaller impact on complexity. Therefore we used $\lambda=0.05$ for our experiments.
With $\lambda = 0$ the complexity is not considered and the equations overfit on the observations: the $R^2$ score tends to improve slightly at the cost of a large increase in complexity, with expressions composed of many different terms attempting to fit perfectly the training set. As the value for the parameter $\lambda$ increases, both the $R^2$ score and the complexity tend to decrease, resulting in equations that underfit the data, as they do not have enough terms to properly capture all the observed dependencies. These results align with \citet{shojaee2023transformerbased}, who introduced the fitness function we use. They chose 0.1 as the final parameter value, which we find performing similarly to 0.05, but slightly underfitting on some benchmarks, particularly R. 

\paragraph{Optimization Loop.} The results suggest that the seed functions generation step plays a key role in our approach, as with $n_s=10$ the results already show a high fitness on the test set, although they still underperform the full method in terms of both $R^2$ and complexity. We notice that using the best seed functions without refinement can outperform the results with ICSR for some values of $\lambda$ (e.g. $\lambda = 0, 0.1$) in the most complex benchmarks (R and Keijzer). This is because the performance is reported on the set of test points and can decrease when refining, due to overfitting on the training points. It's also worth noting that some of the experiments with only a single initial call did not result in any valid seed functions, showing the need for repeating the generation process multiple times. In the prompt used to generate the seed functions (reported in Appendix~\ref{app:prompts}) we specifically ask for a diverse and complex set of functions that can be optimized, which is likely why the complexity on the seed functions is much higher, as it will be lowered later in the optimization loop. Overall, both parts of the method are necessary for the best possible performance; repeating the seed function generation step multiple times allows the model to generate a large number of potential initial expressions, resulting in a solid set of initial candidates for the optimization loop to build upon.

\paragraph{Random Guessing.} As some of the benchmarks contain common equations such as simple polynomials, the LLM could simply be randomly generating functions that fit the data points, instead of actually making use of the information provided in the prompt. To ensure that this is not the case, we compare ICSR with a 'Random Guessing' baseline, where the LLM was prompted for 60 times (matching the budget used for ICSR, which uses 10 prompts to generate the seed functions and 50 prompts for the optimization loop) to generate five random functions, without any information about the observations or previous guesses (the prompt is reported in Appendix~\ref{app:prompts}). The results show that this baseline underperforms ICSR on all four benchmarks, especially on Keijzer, the hardest one. Empirically, we observe that the functions generated by the LLM in this way are all extremely simple, mostly constrained to basic polynomials. This confirms that LLMs are able to extract patterns from the prompt and are not simply randomly generating the solutions.

\section{Discussion}
\label{sec:disc}

\paragraph{Optimizing for out of distribution.} A general framework for optimizing the out of distribution performance of a predictive model (such as a symbolic equation) is to regularise its complexity, following the Occam's Razor principle that simpler explanations are preferable to more complex ones, all other things being equal. In our work we use the working definition of complexity as the number of nodes in the expression tree of an equation. However, more optimal choices could be available: for instance, equations containing expressions not defined on all the real domain (such as logarithms and square roots) could be penalised more, as they could be undefined when extrapolating to larger domains. Knowing in advance the full domain in which an equation is supposed to hold could also greatly improve out of distribution performance by filtering out invalid candidate functions. In the case of ICSR, it could also be leveraged as extra information by the LLM.
Furthermore, we observe that numerous equations that we derive with ICSR have extra terms with very small coefficients (e.g. $\mathcal{O}(10^{-3})$) that do not contribute significantly to the shape of the equation and could be safely suppressed, resulting in expressions with a lower complexity. This could be done by modifying the optimization procedure of the coefficients, to eliminate coefficients under a certain threshold, which would be a hyperparameter of the method.

\paragraph{Vocabulary.} In general, most SR methods are limited to a predefined vocabulary of operators and tokens, while LLMs can virtually explore any possible function and combination. An example of this is with the $x_1^{x_2}$ function in the Nguyen benchmark: in \citet{biggio2021neural}, the authors mention that it is not included in the set of equations that their model can fit, while our approach can recover the exact expression. We also observe a similar trend with the other baselines for this specific expression.
In our prompts (see Appendix~\ref{app:prompts}) we include a vocabulary for the LLM, but this is meant more to guide the LLM into the correct search space and is by no means a hard restriction: for example, we observe that ICSR can produce the \texttt{erf} function even if it wasn't reported in this list. Furthermore, any function that can be found in the model's pre-training corpus (fundamentally the Internet) can be potentially added to the prompt at any time if desired, which is impossible for other fixed-vocabulary methods.

\subsection{Limitations}
\label{sec:disc:limit}
Although promising, the approach presented in this work still suffers from some key limitations that hold back its potential as a full Symbolic Regression method. 

\paragraph{Size of the context window.} 
LLMs are provided with a context window, which represents the maximum number of tokens they can process as input at the same time. For instance, Llama3, used for ICSR, has an 8k token context window.
This limits the amount of information that we can include in the prompt, in terms of training datapoints and previously attempted functions with their errors. However, with context-window size increasing, commercially available LLMs like GPT-4 Turbo \citep{openai2023gpt4} and Claude 3 \citep{claude3}, which process over 100k tokens, this issue is likely to be alleviated or lifted completely.

\paragraph{What to include in the prompt?} Including all needed information in the prompt might not be enough, as some research suggests LLMs cannot fully utilize extended contexts \citep{liu2024lost}. In practice, we observe that when too many points are included, the model often continues generating points, especially with two-dimensional functions. Limiting training points in the prompt to 40 (chosen empirically) helps, while all input points are still used for coefficient optimization. 
Some directions to help the model leveraging the information in the data could be to sample the most informative subset of points to fit in the prompt, or present the LLM with higher-level descriptions of the points, rather than feeding them directly to the model. Finally, we hypothesize that presenting the data in different modalities, such as images of the points and plots of the functions, by using multimodal foundation models, might be helpful to incorporate all information available. We experimented with Vision-Language Models, but our attempts in that direction, reported in Section~\ref{app:vision} of the Appendix, were not fruitful so far.

\paragraph{Dimensionality.} 
Using an LLM for higher dimensional inputs is possible, but dimensionality exacerbates the issues presented above. As the number of variables grows, so does the space dedicated to the input points in the prompt, which will naturally confuse the model and obfuscate the structure in the datapoints even further. Specifically fine-tuning an LLM on this kind of examples might show some improvement, but scaling this approach for higher dimensional problems remains a challenge.

\section{Conclusion}
\label{sec:conc}

We show that LLMs paired with the ICSR approach are able to perform Symbolic Regression tasks on classical SR benchmarks. The proposed method matches or outperforms a variety of established SR baselines, while producing simpler expressions that more closely resemble the complexity of the ground truth equations and result in better out of distribution performance.
This work exposes yet another task that LLMs can be leveraged for, thanks to specialized techniques such as ICSR, and shows promise for integrating these models with mathematical reasoning methods. 

\subsection{Future Work}
\label{sec:conc:fut}

As this is one of the first works published on this topic, much work remains to be done.
LLMs allow the inclusion of domain-specific natural language information into the prompt, as explored by \citet{shojaee2024llmsr}. The natural language interface could be further exploited by employing explicit Chain of Thought-like \citep{wei2022chain, kojima2022large} techniques, allowing the model to output even more well-informed guesses at every step and resulting in an interpretable method. Another interesting direction would be to consider tree-based search algorithms on top of the LLM, analogously to the TPSR \citep{shojaee2023transformerbased} approach. As our work proves the intrinsic ability of LLMs to perform SR without taking into consideration any additional inputs, we have hope that future work can build upon ICSR to further leverage foundation models for SR.

\section*{Acknowledgements}
We are grateful to Alexander Ilin and Alberto Zabeo for the fruitful discussions. We thank Aalto-IT (IT Services of Aalto University, Finland) for provided support with computational resources. This work was supported by the Research Council of Finland (Flagship programme: Finnish Center for Artificial Intelligence FCAI, and grants 352986, 358246) and EU (H2020 grant 101016775 and NextGenerationEU).

\bibliography{custom}

\newpage
\clearpage

\appendix

\justifying
\label{app}

\section{Vision-Language Models} 
\label{app:vision}
In this section we report our findings on extending ICSR to Vision-Language Models (VLMs), which we considered a promising direction, but was not successful experimentally, at least with the VLMs that we considered.

\subsection{Vision-Language Extension}
\label{sec:meth:vis}

\begin{figure}[!h]
    \centering
    \begin{subfigure}{0.48\linewidth}
        \includegraphics[width=1.0\textwidth,keepaspectratio]{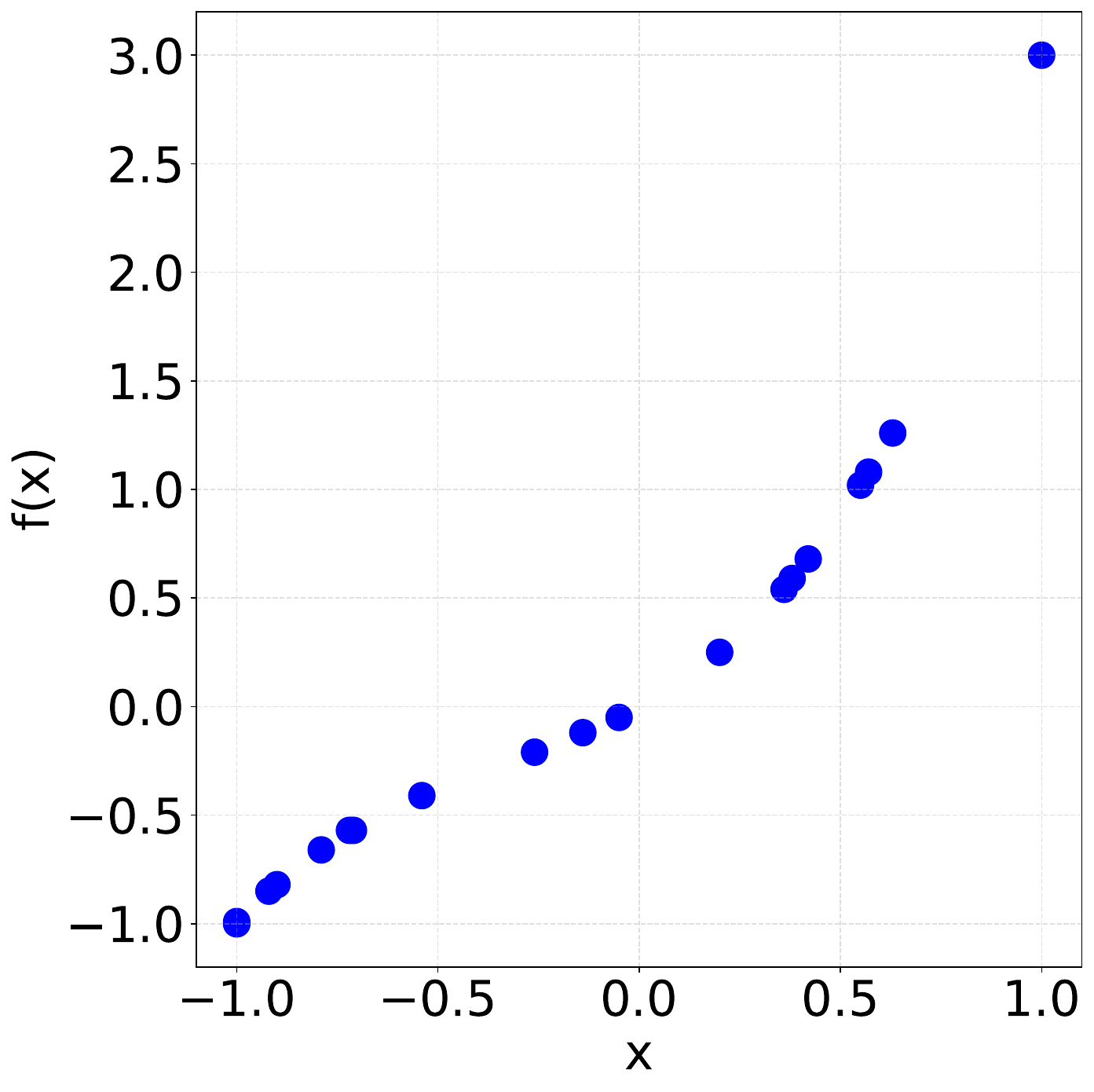} 
        \caption{The observations.}
        \label{fig:points}
    \end{subfigure} \begin{subfigure}{0.48\linewidth}
        \includegraphics[width=1.0\textwidth,keepaspectratio]{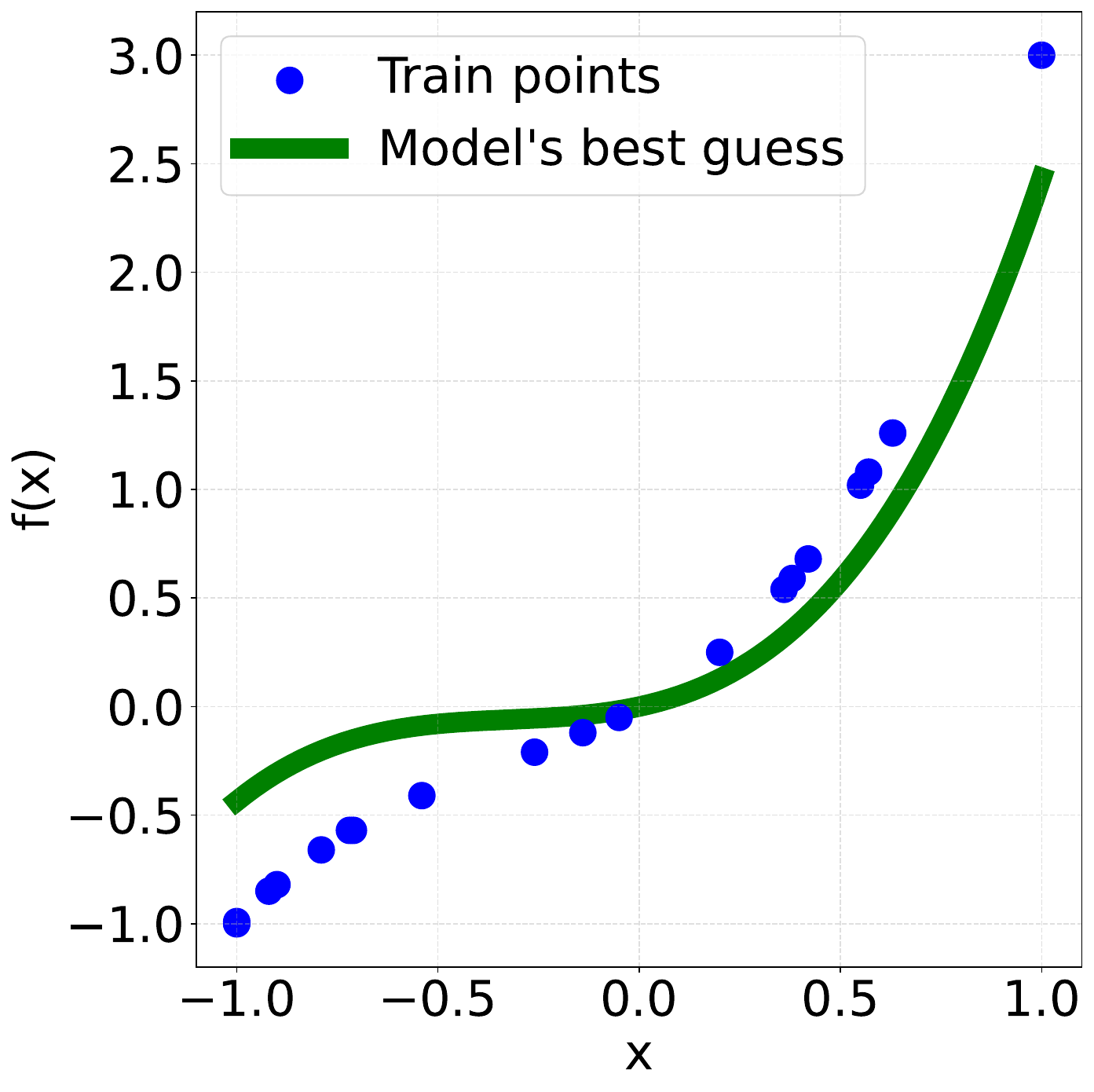}
        \caption{The previous function fit.}
        \label{fig:prev_guess}
    \end{subfigure}
    \caption{\textbf{Example of plots used with the VLM.} (a) Scatter plot of the observations used when generating the seed functions. (b) Plot of the best function from a previous iteration used in the optimization loop.}
    \label{fig:input_plots}
\end{figure}

Reasoning on the observations and the previously attempted functions to come up with better function candidates is a challenging task. Visualising the data and the functions, when possible, can be of great help for humans and, we hypothesize, for SR models too. 
We thus explore the use of visual information in ICSR by considering VLMs in place of LLMs and adding to the optimization meta-prompt a scatter plot containing the observations (Figure~\ref{fig:points}), as well as plots superimposing the best previously generated function (Figure~\ref{fig:prev_guess}). We dub this variant ICSR-V and present results for it in Section~\ref{sec:exp:compar}.
However, the use of both vision and language as input comes with the restriction of dimensionality, as it is impossible to visualize inputs with more than two inputs in a single image. A solution could be to include projections into each dimension as the input, but this can quickly grow out of control as the number of variables increases, and then the additional information would probably provide diminishing returns.

\subsection{Related Work}
VLMs have gained traction after \citet{radford2021clip} introduced CLIP, which aligns text and image representations using a contrastive objective. Various foundation models have been proposed, such as FLAVA \citep{singh2022flava}, LLaVa \citep{liu2023visual, liu2024improved, liu2024llavanext}, Flamingo \citep{alayrac2022flamingo}, OTTER \citep{li2023otter, li2023otterhd}, Fuyu \citep{fuyu-8b} and more recently OpenAI's GPT4's vision extension. A thorough survey of VLM techniques and tasks was performed recently by \citet{zhang2024vision}. Typically, a VLM can be built on top of a pre-trained LLM, which is then paired with an image embedding network that can transfer the image into the same token space used by the model, attempting to keep semantic similarity. This approach is employed, for instance, by BLIP \citep{li2022blip} and its successor BLIP2 \citep{li2023blip2}. Moreover, these models typically can only consume images as input, but are unable to generate them as an answer, but the general framework can be enhanced with methods for text-to-image generation, such as DALL-E \citep{ramesh2021zeroshot, ramesh2022hierarchical} and GILL \citep{koh2023generating}.

\subsection{Comparison of Text-Only and Vision-Language Models}
\label{sec:exp:compar}

To evaluate the effectiveness of the additional plots, we compare our method with a variant using the LLaVa-NeXT \citep{liu2024llavanext} VLM. To ensure a fair comparison, we use the same backbone model and repeat the experiments with and without the inclusion of visual information. This consists of a scatter plot of the observations for the seed functions generation step with the overlay of the best previous function (as the model only supports one input image at the time of writing) during the optimization loop. An example of the input plots can be found in Figure~\ref{fig:input_plots}. We repeat both experiments across five different random seeds and report the results in Table~\ref{tab:vision}. 
Surprisingly, the performance of the method seems to be unaffected by the presence of the images. This might be due to several factors, among which the fact that the vision encoder of the VLM has not been trained on plots of functions, but rather on natural images, thus, the visual inputs might be out of distribution for the model. We also experimented asking the model facts about the plots in input (such as range of points, maximum and minimum values of the function, shape, first and second derivatives), with no consistent success. It might be that future models will be more amenable to this sort of visual mathematical reasoning, but this is not the case for current VLMs, as was also suggested by recent work \citep{wang2024charxivchartinggapsrealistic}.

\begin{table}[]
    \centering
    \resizebox{\linewidth}{!}{
    \begin{tabular}{lcccc}
        \toprule
         \multirow{2}{*}{\textbf{Benchmark}} & \multicolumn{2}{c}{\textbf{ICSR-V}} & \multicolumn{2}{c}{\textbf{ICSR}} \\
         \cmidrule{2-3} \cmidrule{4-5}
          & $R^2$ ($\uparrow$) & $\mathcal{C}$ ($\downarrow$) & $R^2$ ($\uparrow$) & $\mathcal{C}$ ($\downarrow$) \\
        \midrule
        Nguyen & 0.991 $\pm$ 0.003 & 5.1 $\pm$ 0.3 & \textbf{0.994 $\pm$ 0.002} & \textbf{5.0 $\pm$ 0.3} \\
        Constant & \textbf{0.995 $\pm$ 0.001} & 4.3 $\pm$ 0.3 & \textbf{0.995 $\pm$ 0.001} & \textbf{3.9 $\pm$ 0.3} \\
        R & \textbf{0.988 $\pm$ 0.003} & \textbf{5.7 $\pm$ 0.5} & 0.986 $\pm$ 0.003 & \textbf{5.7 $\pm$ 0.4} \\
        Keijzer & \textbf{0.983 $\pm$ 0.006} & 7.6 $\pm$ 0.8 & \textbf{0.984 $\pm$ 0.004} & \textbf{7.4 $\pm$ 0.8} \\
        \midrule
        Overall avg. & 0.989 $\pm$ 0.003 & 5.7 $\pm$ 0.5 & \textbf{0.990 $\pm$ 0.003} & \textbf{5.5 $\pm $ 0.5}\\
        \bottomrule
    \end{tabular}
    } 
    \caption{\textbf{Comparison on the impact of additional visual input.} All experiments are performed with LLaVa-NeXT as the underlying model, either providing or excluding a plot of the best previous function in the prompts (respectively ICSR-V and ICSR columns). We report the averages with their errors.}
    \label{tab:vision}
\end{table}

\section{Hyperparameters}
\label{app:hyper}
We report the hyperparameters used with LLMs (Table~\ref{tab:llm_params}). As reported in the main text, for ICSR we sample $n_s=10$ initial seed functions and repeat the optimization loop for 50 iterations, using an acceptance threshold of 0.99999 and repeating the coefficient fitting for 5 times with different initializations.
For DSR and uDSR we set the computation budget for the number of expressions to evaluate to 200K and extend the vocabulary as \{\texttt{add}, \texttt{sub}, \texttt{mul}, \texttt{div},
\texttt{sin}, \texttt{cos},  \texttt{exp}, \texttt{log}, \texttt{sqrt},
\texttt{n2}, \texttt{abs}, \texttt{n3}, \texttt{n4}\} 
and
\{\texttt{add}, \texttt{sub}, \texttt{mul}, \texttt{div},
\texttt{sin}, \texttt{cos},  \texttt{exp}, \texttt{log}, \texttt{sqrt},
\texttt{abs}, \texttt{poly}\} correspondingly. 
For the NeSymRes model we evaluate the model checkpoint that has been obtained with the training set of 100M expression skeletons. The actual number of the equations in the training set is even larger since the values for the coefficients are resampled on each training batch. The beam size in NeSymRes is set to 10 and the number of restarts for the external coefficient optimizer is 10, while for E2E model the beam size is 100 but the coefficient optimizer is applied just once. E2E doesn't benefit from restarting the external coefficient optimizer as much since E2E predicts the whole equation including the values of the coefficients. The predicted coefficients can be further improved by numerical optimizer but they serve as good initial values.
For all other implementation details, we follow the default hyperparameters provided in the following repositories: gplearn\footnote{\url{https://github.com/trevorstephens/gplearn}}, DSR/uDSR\footnote{\url{https://github.com/dso-org/deep-symbolic-optimization}}, NeSymReS\footnote{\url{https://github.com/SymposiumOrganization/NeuralSymbolicRegressionThatScales}} and E2E/TPSR\footnote{\url{https://github.com/deep-symbolic-mathematics/TPSR}}.

\begin{table}[!h]
    \centering
    \begin{tabular}{cc}
        \toprule
        \textbf{Parameter} & \textbf{Value} \\ \hline
        \texttt{temperature}        & 1.0            \\
        \texttt{top\_p}             & 0.9            \\
        \texttt{top\_k}             & 60             \\
        \texttt{num\_beams}         & 1              \\
        \texttt{max\_new\_tokens}   & 512           \\
        \bottomrule
    \end{tabular}
    \caption{Sampling parameters for the LLMs.}
    \label{tab:llm_params}
\end{table}

\section{Prompts}
\label{app:prompts}

\begin{figure*}[htb]
\centering
\begin{blockquote}
\small
I want you to act as a mathematical function generator.
Given a set of points below, you are to come up with 5 potential functions that would fit the points. Don't worry too much about accuracy: your task is to generate a set of functions that are as diverse as possible, so that they can serve as starting points for further optimization.

To generate the functions, you will start from a set of basic operators and expressions, and combine them into something more complex. 

Your options are:

- An independent variable symbol: x.

- A coefficient symbol: c (there is no need to write a number - write this generic coefficient instead).

- Basic operators: +, -, *, /, \^ , sqrt, exp, log, abs

- Trigonometric expressions: sin, cos, tan, sinh, cosh, tanh

Make sure there are no numbers in the functions, use the coefficient token 'c' instead.
Analyze the points carefully: if there are any negative points in the input, sqrt and log can not be used unless the input is combined with abs.

The functions should all begin with the indicators "f1(x) = ", "f2(x) = "...
Your task is to combine an arbitrary number of these basic blocks to create a complex expression. Don't be afraid to be creative and experiment! The functions should be as complex as possible, combining many different operations. Variety is key!

Points: \{points\}

Functions:

\end{blockquote}
\caption{Prompt used to generate the seed functions.}
\label{prompt:seed}
\end{figure*}

\begin{figure*}[h]
\centering
\begin{blockquote}
\footnotesize
I want you to act as a mathematical function generator.
You are given a set of points with (x, y) coordinates below: 

\{points\}

Below are some previous functions and the error they make on the points above. The errors are arranged in order of their fit values, with the highest values coming first, and lower is better. 

Your task is to give me a list of five new potential functions that are different from all the ones reported below, and have a lower error value than all of the functions below. Only output the new functions and nothing else.

Remember that the functions you generate should always have at most \{num\_variables\} variables \{variables\_list\}. 

The functions should have parametric form, using 'c' in place of any constant or coefficient. The coefficients will be optimized to fit the data. Make absolutely sure that the functions you generate are completely different from the ones already given to you.

The functions should all begin with the indicators "f1(x) = ", "f2(x) = "... 

Remember that you can combine the simple building blocks (operations, constants, variables) in any way you want to generate more complex functions. Don't be afraid to experiment!

\{previous\_trajectory\}
\end{blockquote}
\caption{Prompt used during the optimization loop.}
\label{prompt:opro}
\end{figure*}

\begin{figure*}[h]
\centering
\begin{blockquote}
\footnotesize
Generate five random functions of the form Function: f(x). The functions you generate should always have at most \{num\_variables\} variables \{variables\_list\}. 
Only output the functions and nothing else.
\end{blockquote}
\caption{Prompt used for the random guessing baseline.}
\label{prompt:random}
\end{figure*}

The prompt used to generate the seed functions is reported in Figure~\ref{prompt:seed}, while the prompt used during the optimization loop is reported in Figure~\ref{prompt:opro} and the one used for the random guessing baseline is reported in Figure~\ref{prompt:random}. For the ICSR-V extension presented in Appendix~\ref{app:vision} we add a brief description of the provided plots as well as the image.

\section{Benchmark functions}
\label{app:bench_func}

The list of functions and point ranges for all the benchmarks can be found in Table~\ref{tab:functions}. The range for training and testing points was taken from the original source where available. Nguyen and Constant do not include a range for the testing points, so we used the same range as the training points but with more sample points. $\mathcal{U}$[min, max, num] indicates points randomly sampled from a uniform distribution between the min and max values, while [min, max, num] indicates a range of equispaced points from min to max. The training points are sampled from $\mathcal{U}$[min, max, num] once and then kept fixed across the random seeds and all tested methods to ensure consistency.

\begin{table*}[!t]
    \begin{center}
        \begin{tabular}{cccc}
        \toprule
        \textbf{Experiment} & \textbf{Function} & \textbf{Train Points} & \textbf{Test Points} \\
        \midrule
        nguyen1 & $x^3 + x^2 + x$ & $\mathcal{U}[-1, 1, 20]$ & $[-1, 1, 200]$ \\
        nguyen2 & $x^4 + x^3 + x^2 + x$ & $\mathcal{U}[-1, 1, 20]$ & $[-1, 1, 200]$ \\
        nguyen3 & $x^5 + x^4 + x^3 + x^2 + x$ & $\mathcal{U}[-1, 1, 20]$ & $[-1, 1, 200]$ \\
        nguyen4 & $x^6 + x^5 + x^4 + x^3 + x^2 + x$ & $\mathcal{U}[-1, 1, 20]$ & $[-1, 1, 200]$ \\
        nguyen5 & $\sin(x^2)  \cdot  \cos(x) - 1$ & $\mathcal{U}[-1, 1, 20]$ & $[-1, 1, 200]$ \\
        nguyen6 & $\sin(x) + \sin(x + x^2)$ & $\mathcal{U}[-1, 1, 20]$ & $[-1, 1, 200]$ \\
        nguyen7 & $\log(x + 1) + \log(x^2 + 1)$ & $\mathcal{U}[0, 2, 20]$ & $[0, 2, 200]$ \\
        nguyen8 & $\sqrt{x}$ & $\mathcal{U}[0, 4, 20]$ & $[0, 4, 200]$ \\
        nguyen9 & $\sin(x_1) + \sin(x_2^2)$ & $\mathcal{U}[[-1, -1], [1, 1], 100]$ & $[[-1, -1], [1, 1], 500]$ \\
        nguyen10 & $2 \cdot \sin(x_1) \cdot \cos(x_2)$ & $\mathcal{U}[[-1, -1], [1, 1], 100]$ & $[[-1, -1], [1, 1], 500]$ \\
        nguyen11 & $x_1^{x_2}$ & $\mathcal{U}[[0, 0], [1, 1], 100]$ & $[[0, 0], [1, 1], 500]$ \\
        nguyen12 & $x_1^4 - x_1^3 + \frac{1}{2} \cdot x_2^2 - x_2$ & $\mathcal{U}[[-1, -1], [1, 1], 100]$ & $[[-1, -1], [1, 1], 500]$ \\
        \hline
        constant1 & $3.39x^3 + 2.12x^2 + 1.78x$ & $\mathcal{U}[-1, 1, 20]$ & $[-1, 1, 200]$ \\
        constant2 & $\sin(x^2)  \cdot  \cos(x) - 0.75$ & $\mathcal{U}[-1, 1, 20]$ & $[-1, 1, 200]$ \\
        constant3 & $\sin(1.5x_1) \cdot \cos(0.5x_2)$ & $\mathcal{U}[[-1, -1], [1, 1], 100]$ & $[[-1, -1], [1, 1], 500]$ \\
        constant4 & $2.7x_1^{x_2}$ & $\mathcal{U}[[0, 0], [1, 1], 100]$ & $[[0, 0], [1, 1], 500]$ \\
        constant5 & $\sqrt{1.23x}$ & $\mathcal{U}[0, 4, 20]$ & $[0, 4, 200]$ \\
        constant6 & $x^{0.426}$ & $\mathcal{U}[0, 4, 20]$ & $[0, 4, 200]$ \\
        constant7 & $2\sin(1.3x_1) + \cos(x_2)$ & $\mathcal{U}[[-1, -1], [1, 1], 100]$ & $[[-1, -1], [1, 1], 500]$ \\
        constant8 & $\ln(x+1.4) + \ln(x^2+1.3)$ & $\mathcal{U}[0, 2, 20]$ & $[0, 2, 200]$ \\
        \hline
        keijzer3 & $0.3x  \cdot  \sin(2 \pi x)$ & $\mathcal{{U}}[-1, 1, 100]$ & $[-1, 1, 10000]$ \\
        keijzer4 & \multicolumn{1}{p{5cm}}{$x^3 \cdot \exp(-x) \cdot \cos(x) \sin(x) \cdot (\sin(x)^2 \cdot \cos(x)-1)$} & $[0, 10, 200]$ & $[0.05, 10.05, 200]$ \\
        keijzer6 & $(x \cdot (x+1))/2$ & $\mathcal{{U}}[-1, 1, 50]$ & $[-1, 1, 100]$ \\
        keijzer7 & $\ln(x)$ & $\mathcal{{U}}[1, 100, 100]$ & $[1, 100, 1000]$ \\
        keijzer8 & $\sqrt{x}$ & $\mathcal{{U}}[0, 100, 100]$ & $[0, 100, 1000]$ \\
        keijzer9 & $\ln(x + \sqrt{x^2 + 1})$ & $\mathcal{{U}}[0, 100, 100]$ & $[0, 100, 1000]$ \\
        keijzer10 & $x_1^{x_2}$ & $\mathcal{U}[0, 1, 100]$ & $[0, 1, 1000]$ \\
        keijzer11 & $x_1 \cdot x_2 + \sin((x_1 - 1)  \cdot  (x_2 - 1))$ & $\mathcal{U}[-3, 3, 20]$ & $[-3, 3, 1000]$ \\
        keijzer12 & $x_1^4 - x_1^3 + \frac{(x_2^2)}{2} - x_2$ & $\mathcal{U}[-3, 3, 20]$ & $[-3, 3, 1000]$ \\
        keijzer13 & $6 \cdot \sin(x_1) \cdot \cos(x_2)$ & $\mathcal{U}[-3, 3, 20]$ & $[-3, 3, 1000]$ \\
        keijzer14 & $8/(2+x_1^2+x_2^2)$ & $\mathcal{U}[-3, 3, 20]$ & $[-3, 3, 1000]$ \\
        keijzer15 & $\frac{x_1^3}{5} + \frac{x_2^3}{2}- x_2 - x_1$ & $\mathcal{U}[-3, 3, 20]$ & $[-3, 3, 1000]$ \\
        \hline
        R1 & $(x+1)^3/(x^2-x+1)$ & $\mathcal{U}[-1, 1, 20]$ & $[-1, 1, 20]$ \\
        R2 & $(x^5-3 \cdot x^3+1)/(x^2+1)$ & $\mathcal{U}[-1, 1, 20]$ & $[-1, 1, 20]$ \\
        R3 & $(x^6+x^5)/(x^4+x^3+x^2+x+1)$ & $\mathcal{U}[-1, 1, 20]$ & $[-1, 1, 20]$ \\
        \bottomrule
        \end{tabular}
    \end{center}
    \caption{Functions and point ranges for all benchmarks.}
    \label{tab:functions}
\end{table*}

\section{Sample results}
\label{app:sample}

\begin{figure*}[h]
    \centering
    \includegraphics[width=\textwidth]{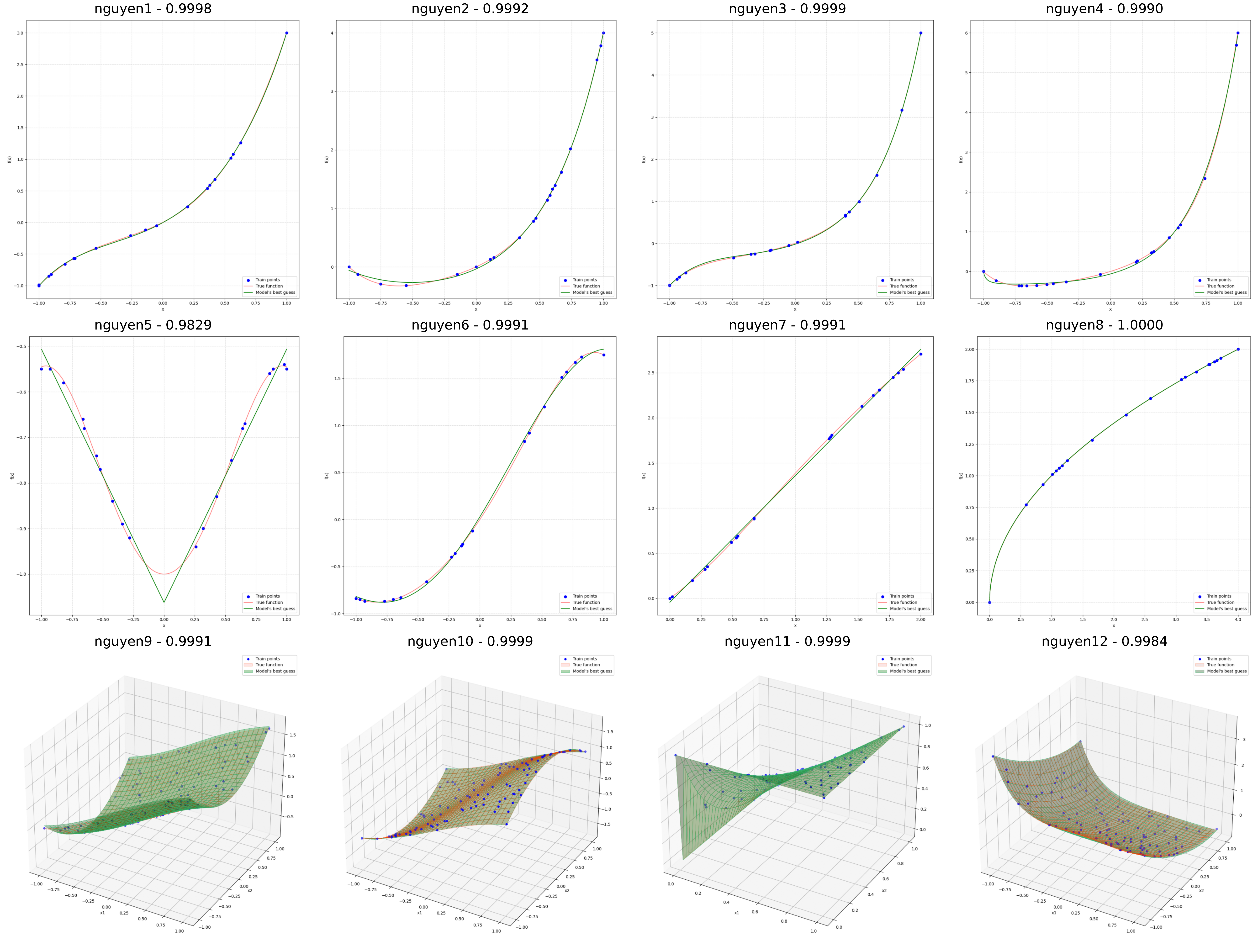}
    \caption{ICSR Results for the Nguyen benchmark for the random seed 1.}
    \label{fig:nguyen_llama}
\end{figure*}

\begin{figure*}[h]
    \centering
    \includegraphics[width=\textwidth]{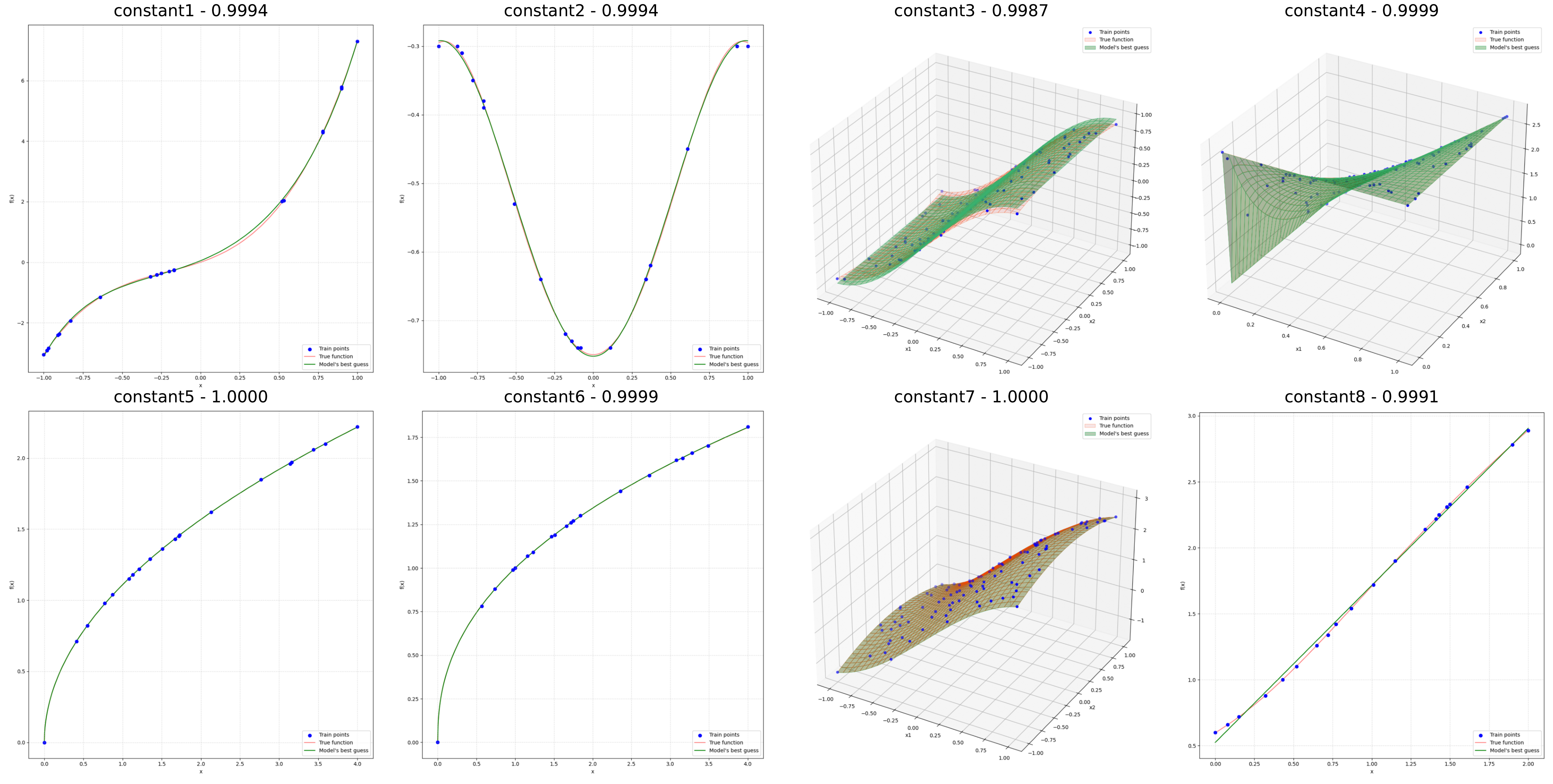}
    \caption{ICSR Results for the Constant benchmark for the random seed 1.}
    \label{fig:constant_llama}
\end{figure*}

\begin{figure*}[h]
    \centering
    \includegraphics[width=0.8\textwidth]{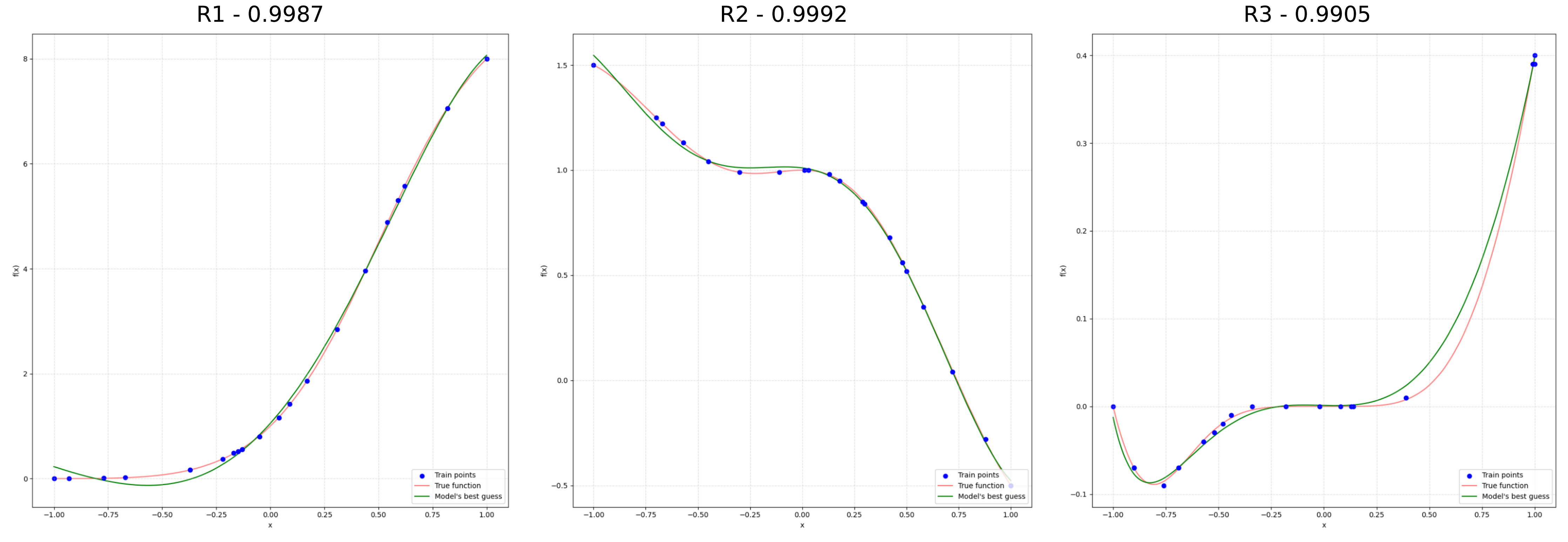}
    \caption{ICSR Results for the R benchmark for the random seed 1.}
    \label{fig:R_llama}
\end{figure*}

\begin{figure*}[h]
    \centering
    \includegraphics[width=\textwidth]{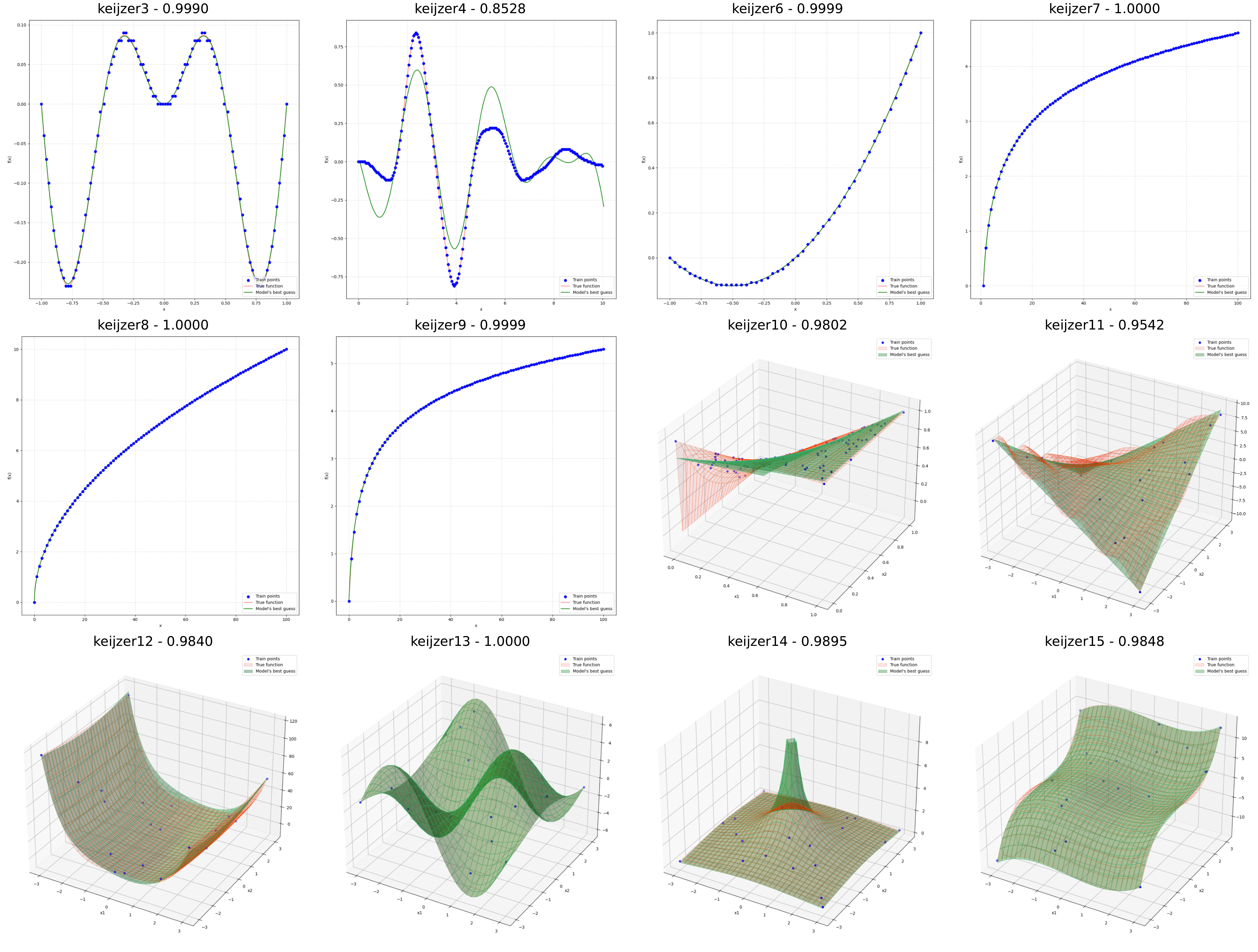}
    \caption{ICSR Results for the Keijzer benchmark for the random seed 1.}
    \label{fig:keijzer_llama}
\end{figure*}

We present a sample of one solution for each function in the benchmarks found by our method, to qualitatively investigate the generated expressions. The observations are seen in blue, the true function is seen in red and the model's guess is seen in green (Figures~\ref{fig:nguyen_llama},~\ref{fig:constant_llama}, and~\ref{fig:R_llama} and~\ref{fig:keijzer_llama}). Some of the failures of the models are apparent: in areas where there is a low density of training points the model sometimes makes guesses that ignore the overall trend, as seen, for example, in the R3 equation (Figure~\ref{fig:R_llama}). The Keijzer benchmark is also much harder in the last 5 equations, with only 20 randomly sampled points to cover a complex 2D space, which can lead to some failures (e.g., in Keijzer 14). 

\end{document}